\documentclass{ws-ijwmip}
\usepackage{graphicx}
\usepackage{cite}
\usepackage{color}
\usepackage{multirow}

\begin{document}

\markboth{Zhuang Qi, Junlin Zhang, Xiaming Chen, Xin Qi}{Comparative Study of Neighbor-based Methods for Local Outlier Detection}

\catchline{}{}{}{}{}

\title{Comparative Study of Neighbor-based Methods for Local Outlier Detection
}

\author{Zhuang Qi
}

\address{School of Software, Shandong University, 1500 ShunHua Road,\\
Jinan, Shandong,
China,
z\_qi@mail.sdu.edu.cn}

\author{Junlin Zhang}

\address{School of Mathematics and Computer, Shantou University, No. 5 Cuifeng Road, \\Shantou, Guangdong, China, 21jlzhang@stu.edu.cn}

\author{Xiaming Chen* (corresponding author)}

\address{School of Mathematics and Computer, Shantou University, No. 5 Cuifeng Road, \\Shantou, Guangdong, China, chenxm@stu.edu.cn}

\author{Xin Qi* (corresponding author)}

\address{School of Chemistry and Life Sciences, Suzhou University of Science and Technology, \\99 XueFu Road, Suzhou, Jiangsu, China, qixin@usts.edu.cn}

\maketitle


\begin{abstract}
The neighbor-based method has become a powerful tool to handle the outlier detection problem, which aims to infer the abnormal degree of the sample based on the compactness of the sample and its neighbors. However, the existing methods commonly focus on designing different processes to locate outliers in the dataset, while the contributions of different types neighbors to outlier detection has not been well discussed. To this end, this paper studies the neighbor in the existing outlier detection algorithms and a taxonomy is introduced, which uses the three-level components of information, neighbor and
methodology to define hybrid methods. This taxonomy can serve as a paradigm where a novel neighbor-based outlier detection method can be proposed by combining different components in this taxonomy. A large number of comparative experiments
were conducted on synthetic and real-world datasets in terms of performance comparison and case study, and the results show that reverse K-nearest neighbor based methods achieve promising performance and dynamic selection method is suitable for working in high-dimensional space. Notably, it is verified that rationally selecting components from this taxonomy may create an algorithms superior to existing methods.
\end{abstract}



\section{Introduction}	

In recent years, with the progress of data acquisition technology, a large number of datasets have been created and applied, which have brought great benefits to scientific research and social development. However, despite the large amount of data that can be used directly, some events are rare or abnormal. Previous studies have shown that the original dataset may contain about 10\% of outliers \cite{hampel2011robust,qi2021iterative}. In many areas, outliers often lead to unreliable results in model decisions. In order to detect outliers, a simple but effective method is to compare the characteristics of the target sample and the samples in its neighborhood. However, the existing outlier detection methods focus on designing different indicators to measure the characteristics of each sample \cite{ha2014robust,tang2017local,wang2020dynamic}, but ignore the significance of neighborhood.

Existing neighbor-based methods can be divided into different categories, based on their approaches to construct the neighbor and the way to select neighbor. Specifically, commonly-used neighbor construct methods include static sorting \cite{ha2014robust, liu2020scalable} and dynamic selection \cite{wang2021local, wang2015fast}. And frequently-used neighbor select approaches can be categorized into K-nearest neighbor (KNN) based  \cite{gao2020cube,ha2014robust,tang2017local,wang2021local}, nature neighbor (NaN) based \cite{huang2016non,wahid2021nanod,zhu2016natural}, reverse K-nearest neighbor (RKNN) based \cite{bryant2017rnn, gao2020cube, radovanovic2014reverse}, Hybrid-based \cite{zhu2016natural} and non-nature neighbor (Non-NaN) based \cite{qi2022novel}. Particularly worth mentioning is the existing methods generally perform well on global outlier problems, while their effectiveness on the task of detecting local outliers has not been fully validated. Although some researchers have compared the scalability, memory consumption and robustness of multiple outlier detection algorithms, the influence of neighbors on the algorithm is not the focus of discussion \cite{domingues2018comparative,xu2018comparison}. To this end, it is necessary to perform a comparative study of existing neighbor-based algorithms to verify their reliability in local outlier detection scenarios.

To analyze and alleviate the aforesaid issues, the basic components of existing neighbor-based methods are decoupled and reorganized, and a taxonomy is proposed for local outlier detection. In this taxonomy, the key components of three levels are applied to define the existing methods, including Information, Methodology and Neighbor levels. The existing methods are segmented into original data and attribute reduction data on data level in detail. At the neighbor level, they are divided into three categories based on the method of neighbor selection, including KNN-based, NaN-based and Non-NaN based. And methodology level includes static sorting and dynamic searching. The taxonomy can act as a new design paradigm for local outlier detection methods and provide an idea to create novel algorithms by combining the components contained in each level of the taxonomy. A large number of experiments have been implemented on the commonly used outlier detection datasets to compare the performance of existing methods and their variants. By observing the results of comparative experiments, the effectiveness of the combination of various neighbor selection and neighbor construction methods is analyzed, and the law of the influence of various neighbors on the performance of the model is summarized. And the prospects of neighbor-based approaches for improving performance in local outlier problems. In general, the main contributions of this paper can be summarized as follows:
\begin{itemize}
	\item Review the construction and selection methods of various neighbors.  
	\item A taxonomy is presented to categorize the key components of neighbor-based methods for local outliers detection and it is conducive to come up with novel algorithms.
	\item The effectiveness of the existing neighbor-based methods and their variants in the local outliers detection tasks is verified. In addition, the role of components at each level is also discussed, which provides guidelines for the development of neighbor based local outlier detection methods in the future.
\end{itemize}

The remaining paper is organized as follows. In Section 2, we introduce quiet a few applications of neighbor-based methods in classifiction and the main ideas of different types of outlier detection algorithms. In Section 3, the taxonomy of neighbor-based methods for local outlier detection is described from three levels: information, methodology, and neighbor. Section 4 details the evaluation metrics, experimental data, experimental results and corresponding analysis. Finally, some conclusions are summarized in Section 5.

\section{Related work}
\subsection{Neighbor-based Methods in classification}

The neighbor-based approaches have attracted a lot of attention for their simplicity and effectiveness, which are widely used in classification tasks \cite{gallego2022efficient,rastin2021generalized,zhang2020cost,zhang2019novel}. Neighbor-based methods often infer the category of the target sample from the characteristics of neighbors, so neighbor feature extraction, neighbor construction and neighbor selection are three key steps of neighbor-based methods \cite{latifi2021dbhc,moutafis2019efficient}. Euclidean distance \cite{liu2018mechanisms,masud2019generate} and Manhattan distance \cite{dewi2019performance} are commonly used indicators to represent features. Neighbors are usually constructed by static sorting and dynamic selection. There are many ways to select neighbors, including K nearest neighbors (KNN), K natural neighbors (NaN), and so on.

\begin{figure}[h]	
	\centering
	\includegraphics[scale=0.5]{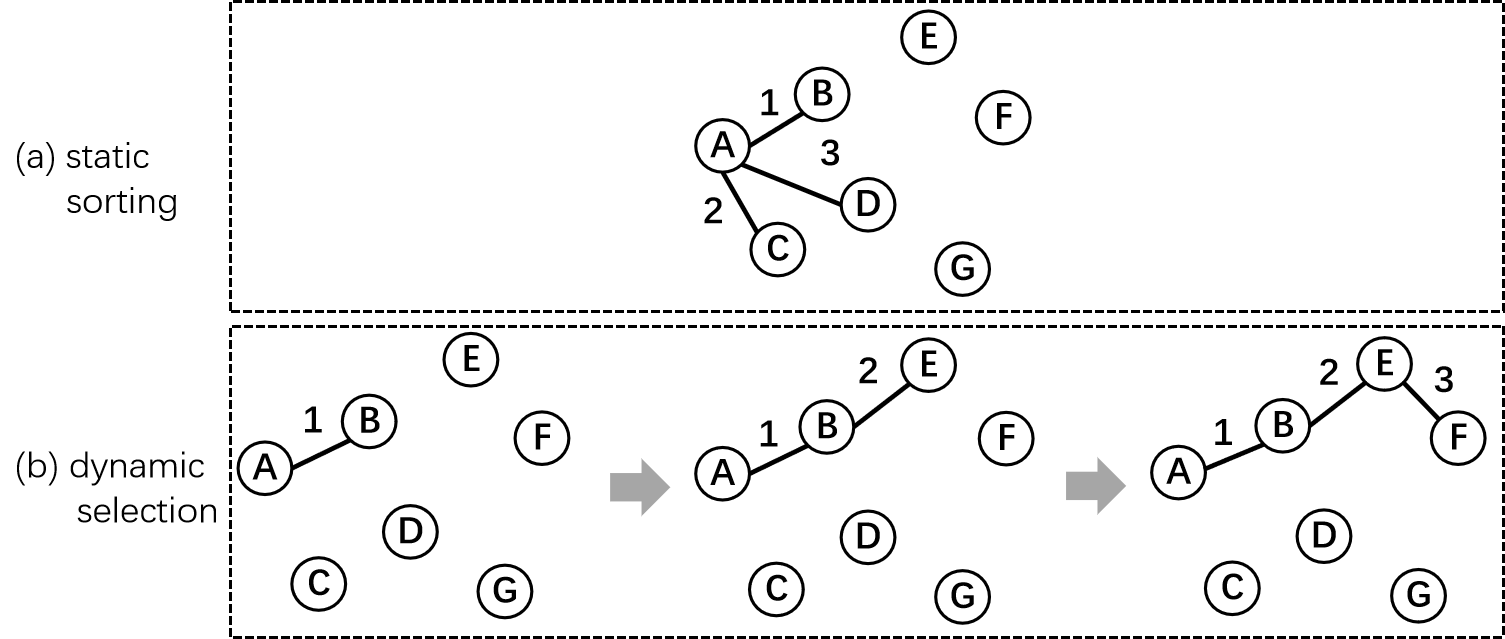}
	\caption{ Compare the neighbors obtained by (a) static sorting and (b) dynamic selection.}
	\label{nn}
\end{figure}

The process of static sorting method is simple and easy to implement, which sorts the distance indicators (or other indicators) between samples, and selects the nearest samples as neighbors \cite{kim2019moving}, as shown on the left in Figure \ref{nn} (a). Dynamic selection is more complicated than static sorting, which only one neighbor is selected in each round. Applying dynamic selection method to find neighbors is based on the set of selected neighbors rather than a fixed sample. Therefore, the starting point of each round is dynamically changed, as shown on the right in Figure \ref{nn} (b). The KNN set obtained by static sorting always quantifies a circular or spherical local area, which is not flexible enough because it cannot be changed according to the specific distribution of the data. In contrast, the idea of dynamic selection helps to generate the correct local neighborhood to quantify arbitrary distributions because it is dynamic rather than based on fixed objects \cite{wang2021local}.

\subsection{Outlier detection}

Outlier detection approaches are generally divided into distance-based \cite{boukela2022approach,radovanovic2014reverse,zhang2009new}, density-based \cite{tang2017local,riahi2021new,wang2021local}, cluster-based \cite{bryant2017rnn,huang2017novel}, and so on. The distance-based methods detect outliers by calculating the distance between all objects. However, they do not consider the change of local density and perform poorly in local outlier detection tasks, so they can only be applied to global outlier detection scenarios \cite{huang2016non}. In density-based methods, abnormal patterns are described as the huge difference in density between an object and its neighbors, and these methods can address the problem of the uneven density of datasets. In cluster-based methods, an object that does not belong to any cluster, or belongs to a small cluster far from other clusters, is identified as an outlier. The purpose of cluster-based methods is to form clustering rather than detect outliers, so outliers can also be called a by-product of clustering.

Neighbors are a key part of the outlier detection algorithm \cite{duggimpudi2019spatio,yepmo2022anomaly}. Due to the unreliability of traditional distance in high-dimensional space, it is necessary to find neighbors in a subspace. In \cite{zhao2017loma}, an outlier detection algorithm based on subspace is proposed, which deletes irrelevant attributes and samples from datasets by analyzing the correlation of attributes. In \cite{liu2017efficient}, the low-rank approximation technique is adopted, whose purpose is to project high-dimensional space into low-dimension.  Outlier detection in subspace is helpful to improve the performance of the model. In addition, the choice of hyperparameter $K$ is also crucial. In \cite{ning2018parameter}, a parameter K search algorithm based on mutual neighbor graph (MNG) was proposed to determine parameter K by looking for the stable state of MNG.

\begin{figure}[t]	
	\centering
	\includegraphics[scale=0.4]{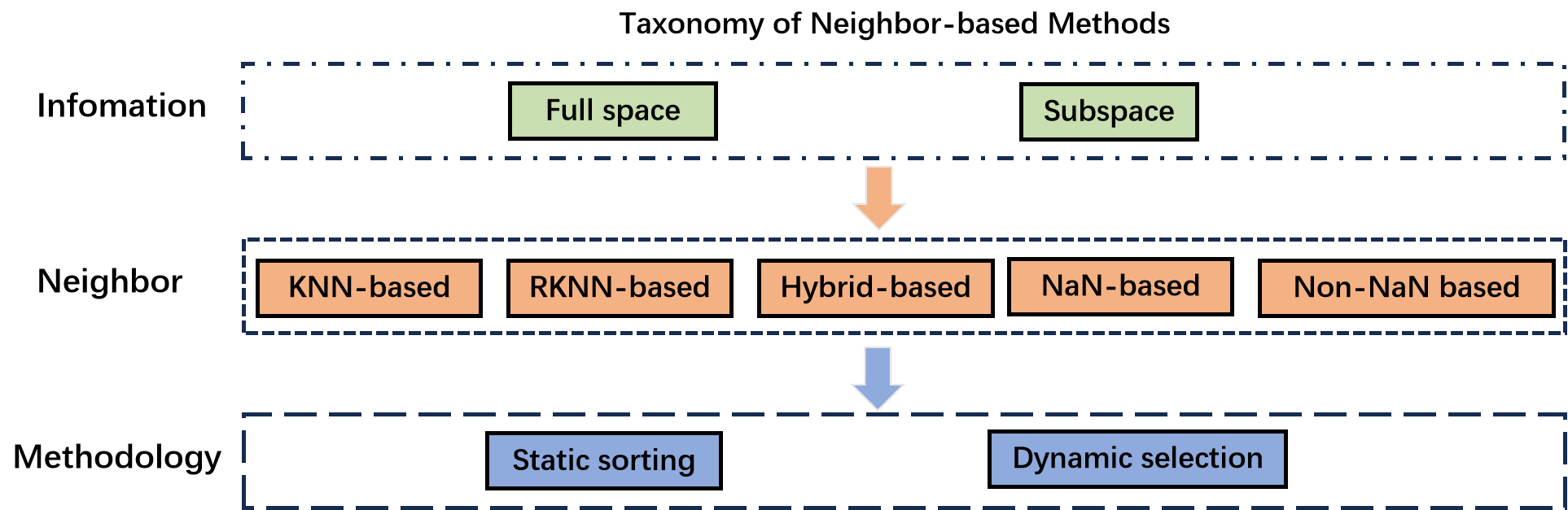}
	\caption{An introduction of the taxonomy for neighbor-based methods in outlier detection.}
	\label{Taxonomy}
\end{figure}

\section{Taxonomy of Neighbor-based Methods for Local Outlier Detection}
The proposed taxonomy decouples and recombines components at different levels (including information, methods, and optimizations, as shown in Figure \ref{Taxonomy}) to describe existing neighbor-based approaches in local outlier detection. The following sections describe them in detail.
\subsection{Information level}
At the information level, the existing neighbor-based methods are divided into two categories: full space-based and subspace-based, as shown in the following.

\subsubsection{Full space-based method}
\begin{figure}[t]
	\centering
	\includegraphics[scale=0.4]{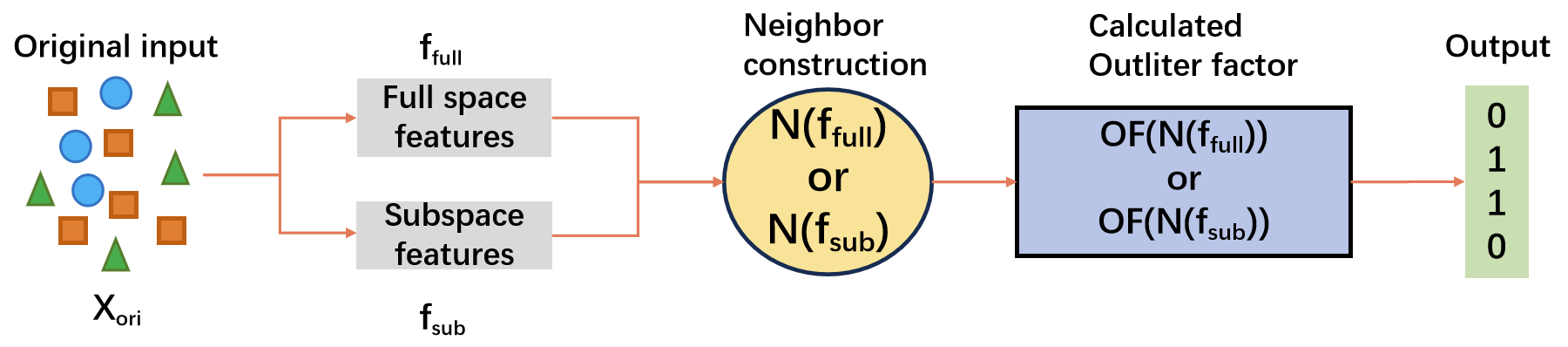}
	\caption{Illustration of the detection process of the neighbor-based outlier detection methods.}
	\label{detection}
\end{figure}

The full space-based methods directly use the original data without data processing. As shown in the Figure \ref{detection}, we summarize the process of neighbor-based outlier detection.

The full space-based methods ignore the second step. Existing works \cite{ha2014robust,liu2020scalable,tang2017local} detect outliers in the full space by designing different outlier factors, which has a good performance in low-dimensional datasets.

\subsubsection{Subspace-based method}
The subspace-based method maps the high-dimensional space of the original data to the low-dimensional space to ensure the reliability of the indicator (such as euclidean distance) \cite{leng2011outliers}. The general form can be expressed as the following formula (1):

\begin{equation}\label{eq1}
	X_{sub}  = S(X_{ori} )
\end{equation}

where $S( \cdot )$ can be expressed as different subspace searching methods, $X_{ori}$ is raw data and $X_{sub}$ is the corresponding subspace data. The transformation from high dimension to low dimension is realized by feature extraction \cite{pascoal2012robust} and feature selection \cite{kriegel2009outlier,yang2011finding}.

\subsection{Methodology level}
Methodology level contains the existing methods of constructing neighbor. Static sorting methods select the K nearest neighbor directly, while dynamic selection methods can only select one (or multiple with the same distance) neighbor in each round until there are K neighbors.

\subsubsection{Static sorting}
Firstly, the distance between each sample is calculated based on the sample features. After sorting, the nearest K objects are selected as KNN set at once. Figure \ref{nn} (a) describes the method of static sorting. For A, the nearest one is C, followed by B, E, F, H, D and G. Therefore, the KNN set of A is $\{ {\rm{C, B, E, F}}\}$ when $K=4$. Changing the K value, the corresponding KNN set can also be obtained quickly. This method is widely used in existing studys owing to the simplicity \cite{bhattacharya2015outlier,ha2014robust,huang2016non,tang2017local,wang2020dynamic}.

\subsubsection{Dynamic selection}
As shown in Figure \ref{nn} (b), after multiple selections, KNN set is obtained, which is different from the result obtained by the static sorting method. From A to B, there are four iterations, and only consider the nearest neighbor or neighbors pointing to the subset of KNN in each iteration.

Obviously, from the perspective of the human social networks, the dynamic selection approaches consider friends of a person as well as his or her friends when constructing KNN sets \cite{wang2021local}. And the KNN set of two people who are the best friend with each other is very similar. 

\begin{figure}[h]
	\centering
	\includegraphics[scale=0.4]{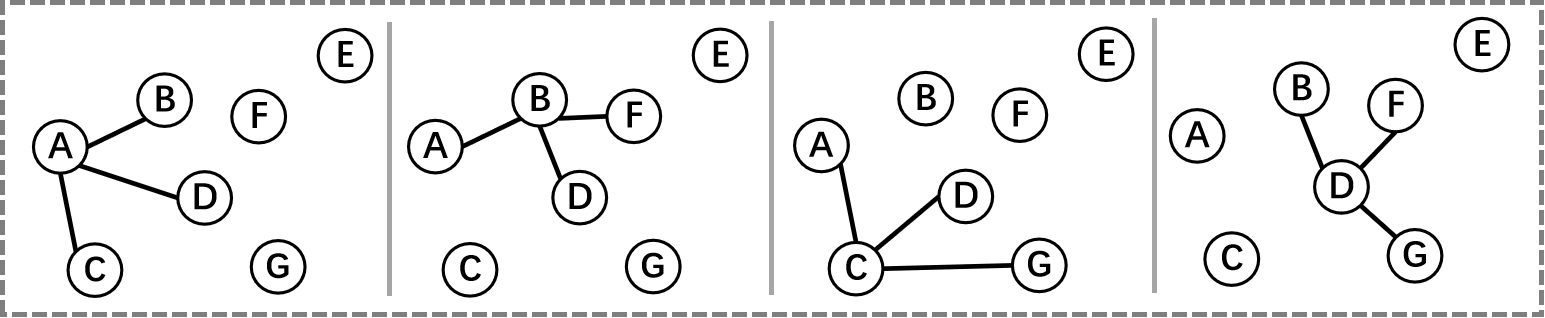}
	\caption{From left to right, the KNN set of A, B, C, and D are displayed respectively.}
	\label{neighbor}
\end{figure}

\subsection{Neighbor level}
Neighbor level divides neighbors into five categories based on different selection methods of neighbors.
We have shown them in Figure \ref{Taxonomy}: which are (1) KNN-based, (2) RKNN-based, (3) Hybrid-based (4) NaN-based, (5) Non-NaN based.
Among them, NaN set and Non-NaN set are constructed based on KNN set and RKNN set. Next, we will introduce the specific method from the details.

\subsubsection{KNN-based}
The way to generate KNN set can be defined as formula \eqref{eq2},
\begin{equation}\label{eq2}
	KNN(x) = N_k (x)
\end{equation}
where $N(\cdot )$ is neighbor selection method, static sorting or dynamic selection, and the corner mark $k$ indicates the number of neighbors.
\subsubsection{RKNN-based}
The RKNN set corresponding to a sample $x$ can be expressed as the following set:
\begin{equation}\label{eq3}
	RKNN(x) = \left\{ {z\left| {x \in KNN(z)} \right.} \right\}
\end{equation}
If KNN set is regarded as a group of one's friends, then RKNN set can be regards as a team who counted the one as friend.

\subsubsection{Hybrid-based}
The Hybrid set extends KNN set and RKNN set by merging RKNN set on the basis of KNN set. The Hybrid set has the following form, 
\begin{equation}
	Hybrid(x) = \{ KNN(x) \cup RKNN(x)\} 
\end{equation}
The hybrid approaches consider not only one's friends, but also the people who regard him or her as friends.
\subsubsection{NaN-based}
NaN set is obtained based on KNN set and RKNN set, i.e.,
\begin{equation}\label{eq4}
	NaN_k(x) = \left\{ {z\left| {z \in KNN(x) \ and \ z \in RKNN(x)} \right.} \right\}
\end{equation}

Similarly, in human social network, a person’s real friends should be a friend who regards himself as a friend \cite{huang2016non,zhu2016natural}. Therefore, NaN represents a group of real friends in a social network.

\subsubsection{Non-NaN based}
The Non-NaN of an object can be defined as the difference set of KNN and NaN, i.e., 
\begin{equation}\label{eq5}
	Non - NaN(x) = KNN(x) - NaN_k (x)
\end{equation}
Contrary to NaN, Non-NaN is a set of fake friends. Notably, applying Non-NaN based neighbors has not been investigated so far.

Next, we give an example of four neighbors. As is shown in the Figure \ref{neighbor}, from left to right, the KNN set of A, B, C, and D are displayed respectively. Obviously, we have following resullt (K=3), $KNN(A) = \{ B,C,D\}$, $KNN(B) = \{ A,D,F\}$, $KNN(C) = \{ A,D,G\}$, $KNN(D) = \{ B,F,G\}$, $RKNN(A) = \{ B,C\}$, $NaN_3 (A) = \{ B,C\}$, $Non - NaN_3  = \{ D\}$.

\section{Experimental evaluation} \label{sec4}

\subsection{Metrics}
Due to the sparsity of outliers, the dataset presents a class imbalance scenario. Since a high performance classifier can be obtained by identifying all samples as inliers, accuracy cannot be used to evaluate the performance of the classifier. To this end, we choose the AUC value as the main indicator, which is the area under the receiver operating characteristic (ROC) curve. The ROC curves are drawn based on False Positive Rate (FPR) and True Positive Rate (TPR), where FPR and TPR are defined as formula \eqref{eq6} and \eqref{eq7},
\begin{equation} \label{eq7}
	FPR = \frac{{FP}}{{FP + TN}}
\end{equation}
\begin{equation} \label{eq8}
    TPR = \frac{{TP}}{{TP + FN}}
\end{equation}
True Positive(TP) denotes the number of true outliers being predicted as outliers, False Positive (FP) be the number of inliers are classified as outliers, True Negative (TN) means the number of inliers are predicted correctly, and False Negative (FN) denotes the number of outliers are classified wrongly.

In addition, to better illustrate the performance of various algorithms, the detection accuracy of outliers is also shown in the results, in which n and 2n with the highest outlier factor are regarded as outliers respectively (n is the number of outliers in the dataset), i.e.
\begin{equation} \label{eq9}
	Acc_n = \frac{{P_n}}{{R_n}}
\end{equation}
\begin{equation} \label{eq10}
	Acc_{2n} = \frac{{P_{2n}}}{{R_n}}
\end{equation} 
where $P_n$ and $P_{2n}$ represent the number of outliers correctly predicted in the data points of n and 2n with the highest outlier factor, respectively.

\subsection{Datasets}
\begin{figure}[h]
	\centering
	\includegraphics[scale=0.3]{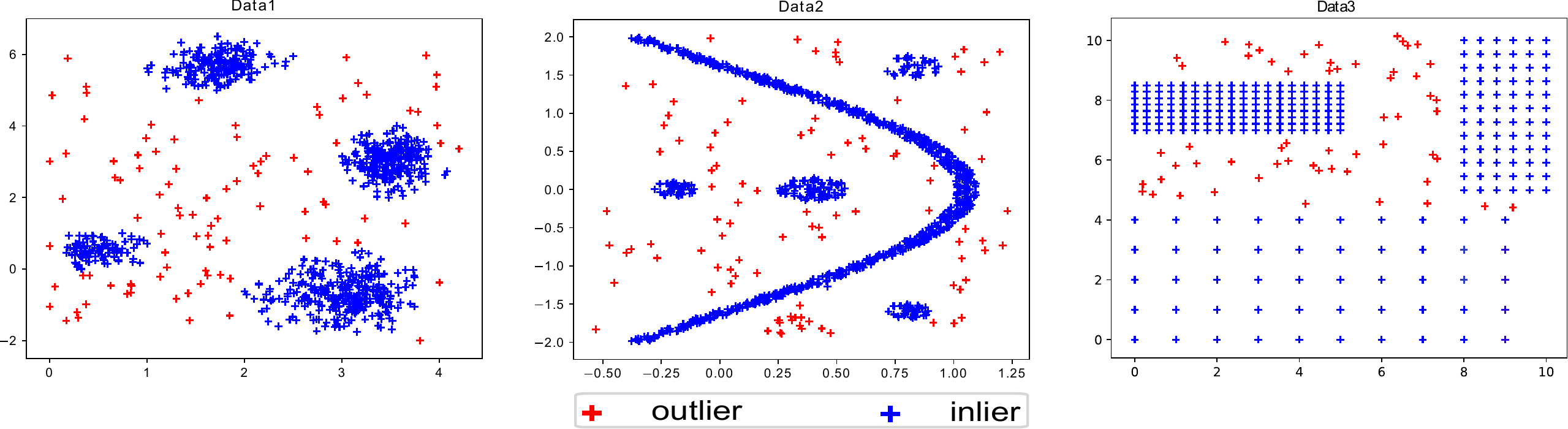}
	\caption{Three synthetic datasets with different clusters. The red '\textcolor{red}{+}' is outliers and the blue '\textcolor{blue}{+}' is inliers.}
	\label{data}
\end{figure}

We evaluate the performance of the existing methods and their variants based on synthetic datasets and real-world datasets. In order to facilitate visualization, we design synthetic datasets in two-dimensional space, and we take into full consideration the diversified clustering patterns, various clustering densities and different clustering sizes when designing the datasets. The details of the synthetic datasets are illustrated in Figure \ref{data}. Similar to paper \cite{ha2014robust,huang2017novel}, Data1 has four clusters with different density and size as inliers, the remaining samples around the cluster are randomly generated as outliers. The total of 940 objects in Data1, and 90 objects (9.57\%) are outliers. Data2 consists of two types of clusters containing 1100 samples with 105 outliers, accounting for 9.54\%.  Data3 involves a low density problem. It has 266 inliers and 58 outliers (17.9\%).

In addition, 9 real-world datasets are used in the study. The number of samples ranged from 148 to 9868 and feature numbers ranged from 5 to 258. Table \ref{dataset} gives an overview of datasets information. All of these datasets are publicly available on UCI\cite{bache2013uci}.

\begin{table}[t]
	\centering
	\caption{\centering The information of 11 datasets.}
	\begin{tabular}{lcccc}
		\hline
		\multicolumn{1}{l}{Datasets}     & Size    & No. of attributes  & No. of outliers    & Ratio of outliers  \\ \hline
		Arrhythmia                       & 248          & 259        & 4     & 1.5\%  \\ 
		PenDigits                        & 9,868        & 16         & 20    & 0.2\%  \\
		Annthyroid                       & 7,200        & 21         & 534   & 7.4\%  \\
		Pima                             & 768          & 8          & 268   & 34.9\%  \\
		Glass                            & 214          & 9          & 9     & 4.2\%  \\
		Wilt                             & 4,839        & 5          & 261   & 5.4\%  \\ 
		Cardiotocography                 & 2114         & 21         & 466   & 22.0\%  \\
		Shuttle                          & 1,013        & 9          & 13    & 1.2\%  \\
		Waveform                         & 3,443        & 21         & 100   & 2.9\%  \\
		WDBC                             & 367          & 30         & 10    & 2.7\%  \\
		Lymphography                     & 148          & 18         & 6     & 4.1\%
		\\
		\hline
	\end{tabular}\label{dataset}
\end{table}

\subsection{Implementation Details}
In the experiment, the method of calculating density and outlier factor refers to \cite{breunig2000lof}. The K value ranged from 5 to 50. Since RKNN set, NaN set, and Non-NaN set may be empty sets, the following rules are applied. First, if RKNN set or NaN set of an object is empty, then the object is seen as outlier. Second, if Non-NaN set of an object is empty, then the object is categorized as inlier. In the following experiments, FP-KNN-SS indicates that KNN set is constructed by Static Sorting method in the Full space, SP-RKNN-DS indicates that in the Subspace, the Dynamic Selection method is used to construct the RKNN set, and so on.
\subsection{Performance Comparison}\label{4.4}
\subsubsection{Accuracy analysis}\label{4.4.1}

\begin{figure}[t]
	\centering
	\includegraphics[scale=0.25]{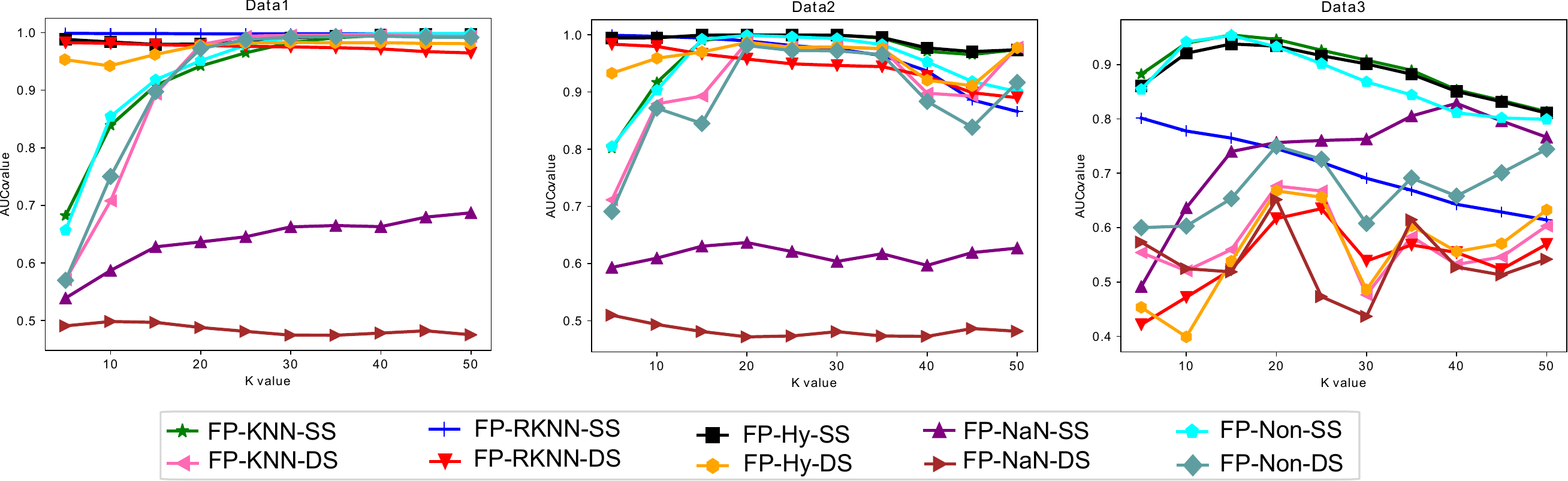}
	\caption{AUC curves of the SS-based algorithms on the synthetic datasets for $5 \le K \le 50$.}
	\label{sdata}
\end{figure}

In this section, we analyze the results of comparative experiments. Firstly, we experiment the whole space method based on different neighborhood and different neighborhood construction methods on synthetic data sets. We set the control parameter k of neighbor size between 5-50. We use different neighborhoods and their construction methods to realize lof \cite{liu2020scalable}. All results are shown in Table \ref{result1}, Figure \ref{sdata}. We can summarize the followings:
\begin{itemize}
	\setlength{\itemsep}{0pt}
	\item In scenarios without low-density patterns, such as Data1 and Data2, RKNN-based methods, better performance than other was achieved. For AUC values, RKNN-based methods get the highest scores. And the outliers in Data2 are identified correctly by FP-RkNN-SS method. It may be caused by the RKNN set contains a lot of data with large density difference from the target sample.  
	
	\item For Data1 and Data2, RKNN-based methods achieve the best Accuracy on top-n and top-2n. This verifies that RKNN can well represent the relationship between samples without low-density patterns. 
	
	\item The RKNN-based has a poor performance on Data3. As shown in Figure \ref{data}, Data3 has three different density clusters. At this moment, Non-NaN based methods achieve the best AUC and Accuracy. This is mainly because some instances in KNN set are  pseudo nearest neighbors, they may have a negative impact on the inference of the model.
	
	\item In Figure \ref{sdata} (right), the blue line decreases with the increase of K value, while the red line fluctuates in the low-level region. This verifies that hyper-parameter K is a key factor to determine the performance of the model.
	
	\item Compared with other methods, KNN-based methods is always neither the best nor the worst. Although FP-KNN-SS method do not perform as well as RKNN-based on Data1 and Data2, most outliers are correctly identified in Data3.
	
	\item From the overall perspective, the method based on static sorting is better than the dynamic selection on synthetic datasets. In two-dimension data, applying static sorting for constructing KNN set can achieve better detections.
	
	\item In addition, the performance of all algorithms has different tendency due to the change of K value. Among them, FP-Hy-SS has been consistent throughout the experiment.
\end{itemize}

\begin{table}[]
	\centering
	\caption{\centering Average AUC values of baseline and its variants for $5 \le K \le 50$ on real-world datasets.}
 \resizebox{0.89\textwidth}{!}{ 
	\begin{tabular}{c|ccc|ccc|ccc}
		\hline
		\multirow{2}{*}{Methods} & \multicolumn{3}{c|}{Data1}                                            & \multicolumn{3}{c|}{Data2}                                            & \multicolumn{3}{c}{Data3}                                            \\ \cline{2-10} 
		& \multicolumn{1}{c|}{$A_n$} & \multicolumn{1}{c|}{$A_{2n}$} & AUC    & \multicolumn{1}{c|}{$A_n$} & \multicolumn{1}{c|}{$A_{2n}$} & AUC    & \multicolumn{1}{c|}{$A_n$} & \multicolumn{1}{c|}{$A_{2n}$} & AUC    \\ \hline
		FP-KNN-SS                & \multicolumn{1}{c|}{0.6822}  & \multicolumn{1}{c|}{0.8066}         & 0.9299 & \multicolumn{1}{c|}{0.8609}  & \multicolumn{1}{c|}{0.9009}         & 0.9610 & \multicolumn{1}{c|}{0.6758}  & \multicolumn{1}{c|}{0.8027}         & 0.8564 \\ 
		FP-RKNN-SS               & \multicolumn{1}{c|}{\textbf{0.9466}}  & \multicolumn{1}{c|}{\textbf{1.0000}}         & \textbf{0.9980} & \multicolumn{1}{c|}{\textbf{0.8438}}  & \multicolumn{1}{c|}{\textbf{0.9276}}         & \textbf{0.9908} & \multicolumn{1}{c|}{0.4586}  & \multicolumn{1}{c|}{0.4517}         & 0.7504 \\ 
		FP-Hy-SS               & \multicolumn{1}{c|}{0.7122}  & \multicolumn{1}{c|}{0.8433}         & 0.9891 & \multicolumn{1}{c|}{0.8223}  & \multicolumn{1}{c|}{0.8980}         & 0.9807 & \multicolumn{1}{c|}{0.6603}  & \multicolumn{1}{c|}{0.7982}         & 0.8545 \\ 
		FP-NaN-SS                & \multicolumn{1}{c|}{0.2335}  & \multicolumn{1}{c|}{0.3122}         & 0.6336 & \multicolumn{1}{c|}{0.3895}  & \multicolumn{1}{c|}{0.4161}         & 0.6154 & \multicolumn{1}{c|}{0.1517}  & \multicolumn{1}{c|}{0.5948}         & 0.7343 \\ 
		FP-Non-SS                & \multicolumn{1}{c|}{0.7377}  & \multicolumn{1}{c|}{0.8466}         & 0.9401 & \multicolumn{1}{c|}{0.7857}  & \multicolumn{1}{c|}{0.8485}         & 0.9443 & \multicolumn{1}{c|}{\textbf{0.7749}}  & \multicolumn{1}{c|}{\textbf{0.8334}}         & \textbf{0.8707} \\ \hline
		FP-KNN-DS                & \multicolumn{1}{c|}{0.7200}  & \multicolumn{1}{c|}{0.7900}         & 0.9118 & \multicolumn{1}{c|}{0.8095}  & \multicolumn{1}{c|}{0.8514}         & 0.9172 & \multicolumn{1}{c|}{0.2068}  & \multicolumn{1}{c|}{0.4620}         & 0.5719 \\ 
		FP-RKNN-DS               & \multicolumn{1}{c|}{\textbf{0.8889}}  & \multicolumn{1}{c|}{\textbf{0.9211}}         & \textbf{0.9784} & \multicolumn{1}{c|}{\textbf{0.8623}}  & \multicolumn{1}{c|}{\textbf{0.9228}}         & \textbf{0.9582} & \multicolumn{1}{c|}{0.0534}  & \multicolumn{1}{c|}{0.1586}         & 0.5421 \\ 
		FP-Hy-DS                 & \multicolumn{1}{c|}{0.7655}  & \multicolumn{1}{c|}{0.8644}         & 0.9756 & \multicolumn{1}{c|}{0.8590}  & \multicolumn{1}{c|}{0.9076}         & 0.9489 & \multicolumn{1}{c|}{0.1810}  & \multicolumn{1}{c|}{0.3810}         & 0.5562 \\ 
		FP-NaN-DS                & \multicolumn{1}{c|}{0.1288}  & \multicolumn{1}{c|}{0.2335}         & 0.4841 & \multicolumn{1}{c|}{0.1742}  & \multicolumn{1}{c|}{0.1838}         & 0.4824 & \multicolumn{1}{c|}{0.0293}  & \multicolumn{1}{c|}{0.0482}         & 0.5374 \\ 
		FP-Non-DS                & \multicolumn{1}{c|}{0.7233}  & \multicolumn{1}{c|}{0.7944}         & 0.9172 & \multicolumn{1}{c|}{0.7257}  & \multicolumn{1}{c|}{0.7914}         & 0.8342 & \multicolumn{1}{c|}{\textbf{0.4491}}  & \multicolumn{1}{c|}{\textbf{0.5206}}         & \textbf{0.6733} \\ \hline
	\end{tabular}} \label{result1}
\end{table}

\begin{table}[]
	\centering
	\caption{\centering Average AUC values of baseline and its variants for $5 \le K \le 50$ on real-world datasets.}
	\resizebox{\linewidth}{!}{
	\begin{tabular}{c|ccccccccccc}
		\hline
		Methods    & Arrhythmia & PenDigits & Annthyroid & Pima  & Glass & WDBC  & Wilt  & Cardiot & Shuttle & Waveform & Lympho \\ \hline
		FP-KNN-SS  & 0.976      & 0.897     & \textbf{0.648}      & 0.634 & 0.756 & 0.977 & \textbf{0.741} & 0.534   & 0.909   & 0.731    & 0.971  \\ 
		FP-RKNN-SS & \textbf{0.986 }     & \textbf{0.986}     & 0.576      & \textbf{0.820} & \textbf{0.819} & \textbf{0.999} & 0.691 & 0.651   & 0.911   & \textbf{0.775}    & \textbf{0.995}  \\ 
		FP-Hy-SS   & 0.977      & 0.966     & 0.649      & 0.688 & 0.794 & 0.998 & 0.735 & 0.557   & \textbf{0.944}   & 0.738    & 0.981  \\ 
		FP-NaN-SS  & 0.728      & 0.537     & 0.569      & 0.601 & 0.207 & 0.806 & 0.632 & 0.532   & 0.473   & 0.543    & 0.634  \\ 
		FP-Non-SS  & 0.977      & 0.916     & 0.638      & 0.655 & 0.794 & 0.978 & 0.733 & \textbf{0.858}   & 0.891   & 0.762    & 0.976  \\ \hline
		FP-KNN-DS  & 0.968      & 0.929     & 0.633      & 0.796 & 0.825 & 0.966 & 0.698 & 0.529   & 0.842   & 0.702    & 0.988  \\ 
		FP-RKNN-DS & 0.960      & 0.936     & 0.597      & \textbf{0.845} &\textbf{ 0.844} & \textbf{0.999} & 0.694 & 0.659   & \textbf{0.921}   & 0.613    & 0.986  \\ 
		FP-Hy-DS   & \textbf{0.971}      & \textbf{0.968}     & \textbf{0.634}      & 0.813 & 0.875 & 0.998 & \textbf{0.713} & 0.552   & 0.868   & \textbf{0.705}    & \textbf{0.991}  \\ 
		FP-NaN-DS  & 0.737      & 0.597     & 0.436      & 0.482 & 0.653 & 0.809 & 0.426 & 0.586   & 0.625   & 0.438    & 0.091  \\ 
		FP-Non-DS  & 0.964      & 0.933     & 0.618      & 0.784 & 0.825 & 0.973 & 0.684 & \textbf{0.726}   & 0.815   & 0.693    & 0.989  \\ \hline
		SP-KNN-SS  & 0.981      & 0.864     & 0.782      & 0.632 & 0.749 & 0.982 & 0.474 & 0.558   & 0.953   & 0.735    & 0.978  \\ 
		SP-RKNN-SS & 0.978      & \textbf{0.988}     & 0.793      & \textbf{0.833} & \textbf{0.825} & \textbf{0.999} & 0.248 & \textbf{0.679}   & 0.929   & \textbf{0.785}    & \textbf{0.985}  \\ 
		SP-Hy-SS   & 0.982      & 0.944     & \textbf{0.840}      & 0.707 & 0.791 & 0.998 & 0.444 & 0.581   & 0.951   & 0.742    & 0.977  \\ 
		SP-NaN-SS  & 0.792      & 0.631     & 0.659      & 0.601 & 0.231 & 0.825 & 0.484 & 0.552   & 0.481   & 0.565    & 0.938  \\ 
		SP-Non-SS  & \textbf{0.986}      & 0.875     & 0.771      & 0.657 & 0.787 & 0.981 & \textbf{0.499} & 0.571   & \textbf{0.959}   & 0.757    & 0.974  \\ \hline
		SP-KNN-DS  & 0.986      & 0.916     & 0.776      & 0.776 & 0.821 & 0.955 & 0.468 & 0.545   & 0.779   & 0.682    & 0.852  \\ 
		SP-RKNN-DS & \textbf{0.998}      & \textbf{0.989}     & 0.775      & \textbf{0.856} & 0.856 & \textbf{0.999} & 0.286 & 0.561   & 0.796   & 0.639    & \textbf{0.953}  \\ 
		SP-Hy-DS   & 0.996      & 0.976     & \textbf{0.789}      & 0.813 & \textbf{0.880} & 0.998 & 0.442  & 0.552   & 0.786   & \textbf{0.684 }   & 0.872  \\ 
		SP-NaN-DS  & 0.842      & 0.606     & 0.389      & 0.474 & 0.617 & 0.876 & 0.481 & 0.543   & 0.809   & 0.463    & 0.508  \\ 
		SP-Non-DS  & 0.992      & 0.921     & 0.747      & 0.774 & 0.832 & 0.962 & \textbf{0.498} & \textbf{0.566}   & \textbf{0.827}   & 0.673    & 0.815  \\ \hline
	\end{tabular}} \label{results2_1}
\end{table}

\begin{table}[]
	\centering
	\caption{\centering Average top-n Accuracy of baseline and its variants for $5 \le K \le 50$ on real-world datasets.}
	\resizebox{\linewidth}{!}{
	\begin{tabular}{c|cccccccccccc}
		\hline
		Methods    & Arrhythmia & PenDigits & Annthyroid & Pima  & Glass & WDBC  & Wilt  & Cardiot & Shuttle & Waveform & Lympho \\ \hline
		FP-KNN-SS  & 0.425      & 0.015     & 0.113      & 0.335 & 0.189 & 0.790 & 0.084 & 0.279   & 0.131   & 0.173    & 0.667  \\ 
		FP-RKNN-SS & \textbf{0.528}      & 0.010     & 0.069      & \textbf{0.552}  & 0.111 & \textbf{0.904} & 0.005 & \textbf{0.322}   & \textbf{0.177}   & \textbf{0.276}    & \textbf{0.750}  \\ 
		FP-Hy-SS   & 0.425      & 0.030     & 0.109      & 0.366 & \textbf{0.199} & 0.870 & 0.049 & 0.295   & 0.131   & 0.177    & 0.683  \\ 
		FP-NaN-SS  & 0.132      & 0.020     & \textbf{0.149}      & 0.259 & 0.022 & 0.200 & \textbf{0.095} & 0.276   & 0.100   & 0.042    & 0.283  \\ 
		FP-Non-SS  & 0.451      & \textbf{0.040}     & 0.111      & 0.396 & 0.156 & 0.800 & 0.074 & 0.144   & 0.131   & 0.228    & 0.733  \\ \hline
		FP-KNN-DS  & 0.675      & 0.005     & 0.127      & 0.495 & 0.089 & 0.800 & 0.016 & 0.245   & 0.069   & 0.127    & 0.617  \\ 
		FP-RKNN-DS & 0.551      & \textbf{0.170}     & 0.104      & \textbf{0.548} & 0.089 & \textbf{0.900} & 0.007 & \textbf{0.362}   & 0.054   & 0.083    & 0.533  \\ 
		FP-Hy-DS   & 0.650      & 0.010     & \textbf{0.146}      & 0.517 & 0.078 & 0.890 & 0.008 & 0.268   & \textbf{0.081}   & \textbf{0.129}    & 0.617  \\ 
		FP-NaN-DS  & 0.212      & 0.011     & 0.055      & 0.219 & 0.111 & 0.720 & \textbf{0.061} & 0.278   & 0.054   & 0.023    & 0.324  \\ 
		FP-Non-DS  & \textbf{0.695}      & 0.023     & 0.092      & 0.468 & \textbf{0.156} & 0.830 & 0.029 & 0.258   & 0.023   & 0.122    & \textbf{0.633}  \\ \hline
		SP-KNN-SS  & \textbf{0.421}      & 0.025     & 0.227      & 0.336 & 0.156 & 0.780 & 0.021 & 0.288   & 0.377   & 0.214    & 0.517  \\ 
		SP-RKNN-SS & 0.125      & 0.005     & \textbf{0.302}      & \textbf{0.562} & 0.111 & \textbf{0.903} & 0.005 & \textbf{0.43}7   & 0.285   & \textbf{0.334}     & 0.55   \\ 
		SP-Hy-SS   & 0.375      & \textbf{0.035}     & 0.250      & 0.395 & 0.156 & 0.860 & 0.016 & 0.297   & 0.385   & 0.219    & 0.500  \\ 
		SP-NaN-SS  & 0.124      & 0.011     & 0.186      & 0.274 & 0.022 & 0.313 & \textbf{0.034} & 0.272   & 0.138   & 0.036    & 0.467  \\ 
		SP-Non-SS  & 0.400      & 0.021     & 0.204      & 0.394  & \textbf{0.194} & 0.820 & 0.015 & 0.311   & \textbf{0.438}   & 0.289    & \textbf{0.534}  \\  \hline
		SP-KNN-DS  & 0.800      & 0.005     & 0.165      & 0.457 & 0.100 & 0.860 & 0.023 & 0.269   & 0.200   & 0.142    & 0.517  \\ 
		SP-RKNN-DS & \textbf{0.950 }      & \textbf{0.145}     & \textbf{0.262}      & \textbf{0.575} & 0.100 & \textbf{1.000} & 0.001 & 0.307   & 0.169   & 0.112    & \textbf{0.750}  \\ 
		SP-Hy-DS   & 0.875      & 0.005     & 0.188      & 0.516 & 0.089 & 0.930 & 0.019 & 0.281   & 0.185   & \textbf{0.148}    & 0.533  \\ 
		SP-NaN-DS  & 0.128      & 0.023     & 0.054      & 0.197 & 0.133 & 0.700 & 0.064 & 0.294   & \textbf{0.446}   & 0.009    & 0.342  \\ 
		SP-Non-DS  & 0.801      & 0.013     & 0.166      & 0.424 & \textbf{0.211} & 0.880 & \textbf{0.072} & \textbf{0.318}   & 0.192   & 0.133    & 0.544  \\ \hline
	\end{tabular}} \label{results2_2}
\end{table}

\begin{table}[]
	\centering
	\caption{\centering Average top-2n Accuracy of baseline and its variants for $5 \le K \le 50$ on real-world datasets.}
	\resizebox{\linewidth}{!}{
	\begin{tabular}{c|ccccccccccc}
		\hline
		Methods    & Arrhythmia & PenDigits & Annthyroid & Pima  & Glass & WDBC  & Wilt  & Cardiot & Shuttle & Waveform & Lympho \\ \hline
		FP-KNN-SS  & 0.675      & 0.080     & 0.246      & 0.528 & 0.200 & 0.890 & 0.211 & 0.483   & 0.415   & 0.269    & 0.833  \\ 
		FP-RKNN-SS & \textbf{0.754}      & 0.020     & 0.127      & \textbf{0.782} & 0.144 & \textbf{1.000 }& 0.013 & \textbf{0.599}   & 0.208   & \textbf{0.354}    & \textbf{1.000}  \\ 
		FP-Hy-SS   & 0.675      & \textbf{0.095}     & 0.239      & 0.601 & 0.200 & \textbf{1.000} & 0.156 & 0.507   & 0.423   & 0.279    & 0.833  \\ 
		FP-NaN-SS  & 0.245      & 0.025     & 0.204      & 0.515 & 0.022 & 0.340 & 0.181  & 0.479   & 0.138   & 0.063    & 0.417  \\ 
		FP-Non-SS  & 0.675      & 0.065     & \textbf{0.244}      & 0.564 & \textbf{0.256} & 0.950 & \textbf{0.215} & 0.256   & \textbf{0.469}   & 0.345    & 0.900  \\  \hline
		FP-KNN-DS  & 0.725      & 0.015     & \textbf{0.195}      & 0.740 & 0.256 & 0.920 & 0.128 & 0.413   & 0.115   & \textbf{0.189}    & 0.983  \\ 
		FP-RKNN-DS & 0.575      & \textbf{0.260}     & 0.169      & \textbf{0.812} & 0.267 & \textbf{1.000} & 0.033 & \textbf{0.632}   & \textbf{0.208}   & 0.145     & \textbf{1.000}  \\ 
		FP-Hy-DS   & \textbf{0.750}       & 0.015     & 0.195      & 0.763 & 0.256 & \textbf{1.000} & 0.092 & 0.427   & 0.108   & 0.185    & \textbf{1.000}  \\ 
		FP-NaN-DS  & 0.018      & 0.012     & 0.073      & 0.384 & \textbf{0.378} & 0.720 & \textbf{0.159} & 0.543   & 0.054   & 0.027    & 0.632  \\ 
		FP-Non-DS  & 0.725      & 0.047     & 0.166      & 0.746 & 0.356 & 0.940 & 0.126 & 0.446   & 0.077   & 0.175    &\textbf{ 1.000}  \\ \hline
		SP-KNN-SS  & 0.775      & 0.035     & 0.433      & 0.570 & 0.233 & 0.910 & 0.068 & 0.477   & \textbf{0.846}   & 0.294    & 0.517  \\ 
		SP-RKNN-SS & 0.650      & 0.025     & 0.459      & \textbf{0.803} & 0.122 & \textbf{1.000} & 0.004 & \textbf{0.625}   & 0.362   & \textbf{0.383}    & \textbf{0.550}  \\ 
		SP-Hy-SS   & 0.775      & \textbf{0.050}    & \textbf{0.501}      & 0.673 & 0.233 & \textbf{1.000} & 0.059 & 0.516   & \textbf{0.846}  & 0.303    & 0.500  \\ 
		SP-NaN-SS  & 0.245      & 0.035     & 0.321      & 0.522 & 0.022 & 0.390 & \textbf{0.095} & 0.502   & 0.169   & 0.075    & 0.467  \\ 
		SP-Non-SS  & \textbf{0.854}      & 0.025     & 0.417      & 0.582 & \textbf{0.256} & 0.940 & 0.084 & 0.511   & 0.838   & 0.354    & 0.517  \\ \hline
		SP-KNN-DS  & 0.900      & 0.067     & 0.410      & 0.714 & 0.311 & 0.900 & 0.065 & 0.494   & 0.338   & \textbf{0.208}    & 0.517  \\ 
		SP-RKNN-DS & \textbf{1.000}      & \textbf{0.210}     & \textbf{0.428}      & \textbf{0.819} & 0.289 & \textbf{1.000} & 0.008 & 0.491   & 0.277   & 0.182    & \textbf{0.750}  \\ 
		SP-Hy-DS   & 0.950      & 0.075     & \textbf{0.428}      & 0.764 & 0.322 & 0.990 & 0.052 & 0.498   & 0.323   & 0.217    & 0.533  \\ 
		SP-NaN-DS  & 0.024      & 0.045     & 0.067      & 0.374 & 0.278 & 0.810 & 0.076 & 0.485   & \textbf{0.562}   & 0.027    & 0.050  \\ 
		SP-Non-DS  & 0.950      & 0.078     & 0.367      & 0.733 & \textbf{0.356} & 0.910 & \textbf{0.149} & \textbf{0.510}   & 0.300   & 0.189    & 0.317  \\ \hline
	\end{tabular}} \label{results2_3}
\end{table}

\begin{figure}[t]
	\centering
	\includegraphics[scale=0.6]{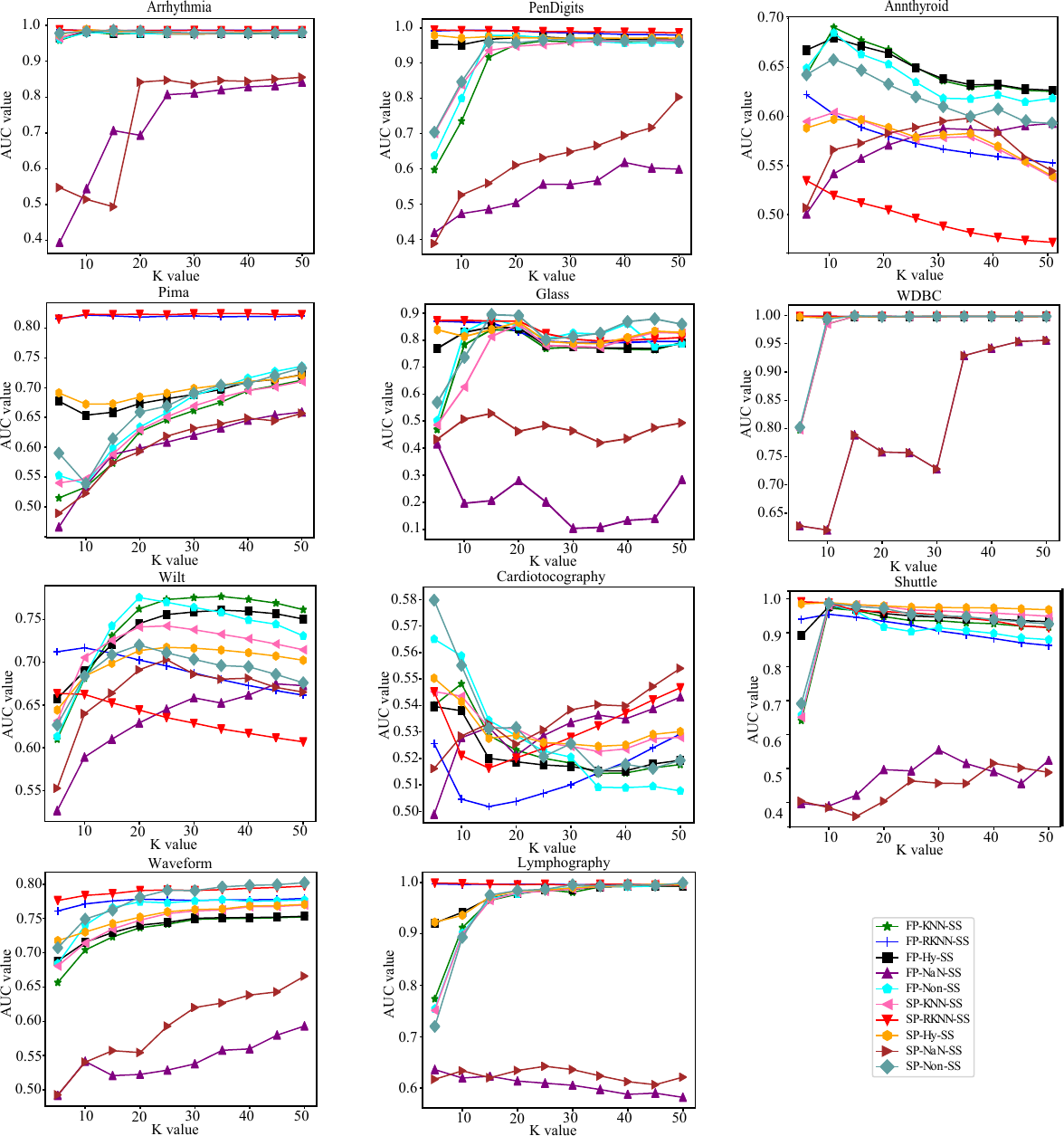}
	\caption{AUC curves of the SS-based algorithms on the real-world datasets for $5 \le K \le 50$.}
	\label{resSS}
\end{figure}

\begin{figure}[t]
	\centering
	\includegraphics[scale=0.6]{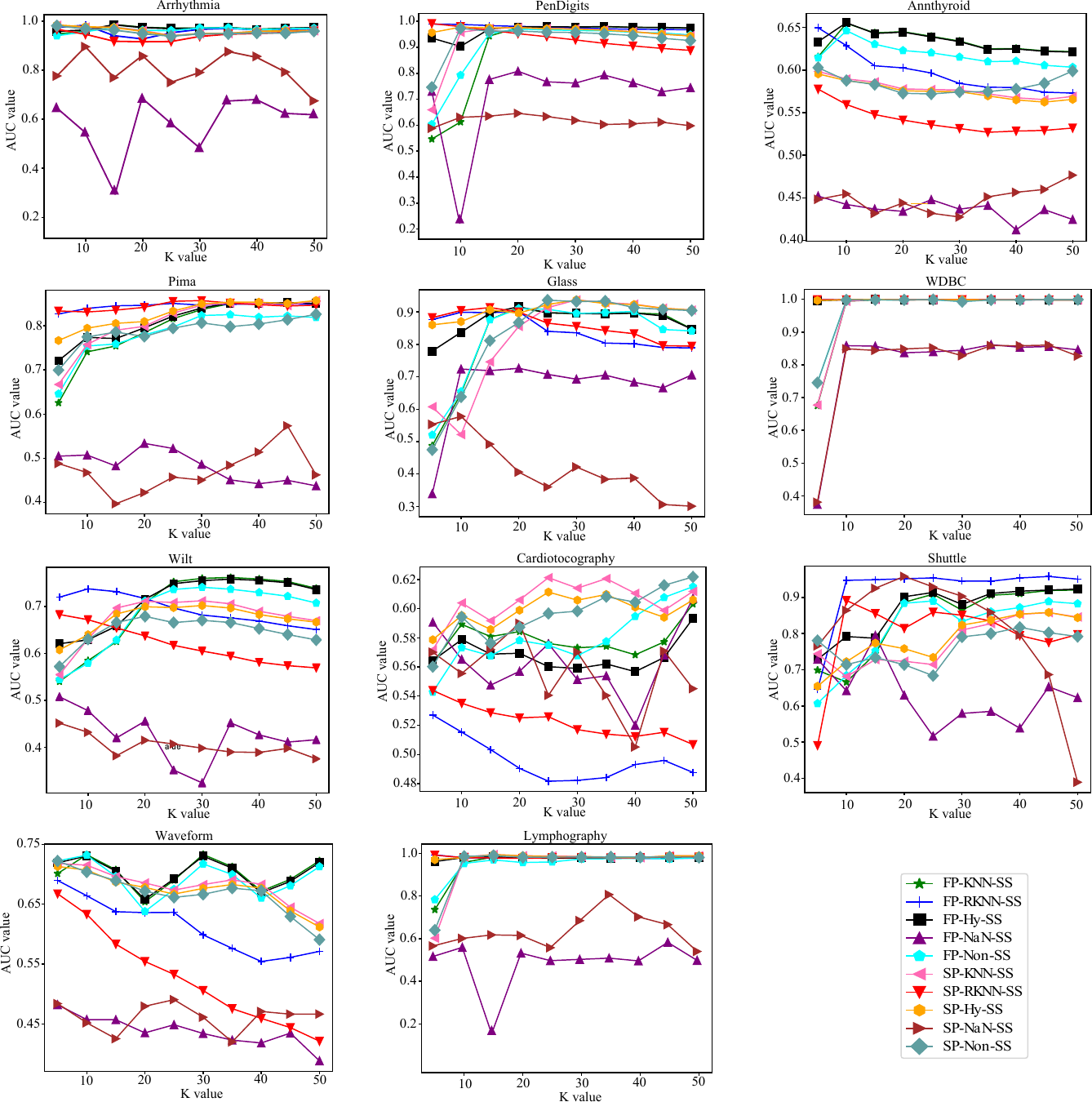}
	\caption{AUC curves of the DS-based algorithms on the real-world datasets for $5 \le K \le 50$.}
	\label{resDS}
\end{figure}

Next, we ran experiments on real-world datasets. Compared with synthetic data, the real-world datasets may contain more samples, the samples have more features, and the data distribution is more complex. In order to solve the problem of unreliable distance in high-dimensional space, PCA method is used to construct subspace and the dimension is reduced to half of the original. The results are reported in Figure \ref{resSS}, \ref{resDS} and Table \ref{results2_1}, \ref{results2_2}, \ref{results2_3}. We have the following points:
\begin{itemize}
	\setlength{\itemsep}{0pt}
	\item In Table \ref{results2_1}, we noticed that RKNN-based method achieved the best performance 22 times in all the comparative experiments. The second is Hybrid-based methods, which got the best 11 times. And compared with other methods, NaN-based methods is always maintains a low level. This result is similar to the above. This is mainly because NaN contains the most similar neighbors, making it impossible to distinguish outliers from inliers.
	
	\item Experiments on most datasets using subspace methods generally outperform full-space methods. In particular, the performance of methods based on dynamic selection and subspace is better than the corresponding full space methods on the Waveform dataset. However, subspace methods sometimes perform poorly, such as they do not perform as well as full space methods on Annthyroid dataset. This shows that constructing a reliable subspace is essential for improving the performance. 
	
	\item As shown in Table \ref{results2_2}, \ref{results2_3}, for top-n and top-2n Accuracy, many methods achieve poor performance on PenDigits and Wilt datasets. They can barely identify outliers. However, this leads a high AUC values. We find an interesting fact that when the proportion of outliers in dataset is small, low accuracy can also lead to a high AUC. With the increase of the proportion of outliers, high AUC corresponds to high accuracy.    
	
	\item From Figure \ref{resSS} and \ref{resDS}, we can learn that the performance of the same algorithm may change greatly under different K values. Among them, the performance of the SP-KNN-SS method on the Glass dataset change the most, where a small K value (K=5) corresponds to an AUC value less than 0.5, while the AUC increases to about 0.9 as the k value increases to 20.
	
	\item Experimental results in some high dimensional datasets show that the method based on dynamic selection is superior to the method based on static sorting. This indicates that methods based on dynamic selection is suitable for high-dimensional data rather than low-dimensional. 
	
	\item The performance of Non-NaN based methods in subspace is stable. Even if performance degrades, it is negligible compared to other methods. For example, on Annthyroid dataset, the AUC value of SP-KNN-SS method decreased by about 0.07 compared with the corresponding full-space method,, but SP-Non-SS decreased by about 0.01.
	
	\item As shown in Figure \ref{resSS} and \ref{resDS}, on both datasets, all methods achieve best performance at or around $K=30$. This can serve as a knowledge guide for us to select the $K$ parameter.
	
\end{itemize}

\subsection{Case Study}\label{4.7}
In this section, we will further analyze the performance of all methods. The detection results of both methods on the synthetic datasets are visualized.  The following observations can be drawn:

\begin{figure}[t]
	\centering
	\includegraphics[scale=0.3]{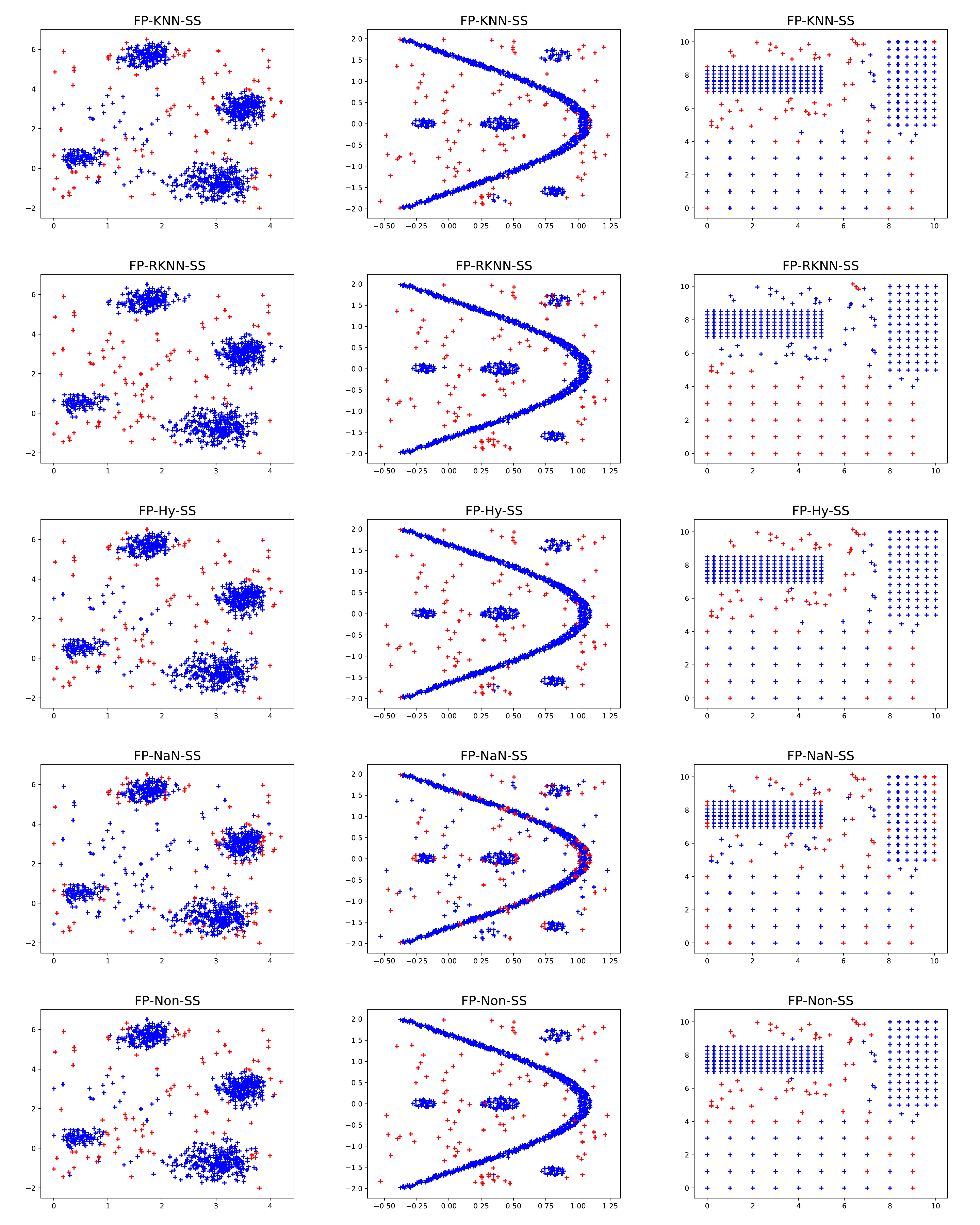}
	\caption{Outlier detection results of the SS-based methods at $K=20$.}
	\label{sss}
\end{figure}

\begin{figure}[t]
	\centering
	\includegraphics[scale=0.3]{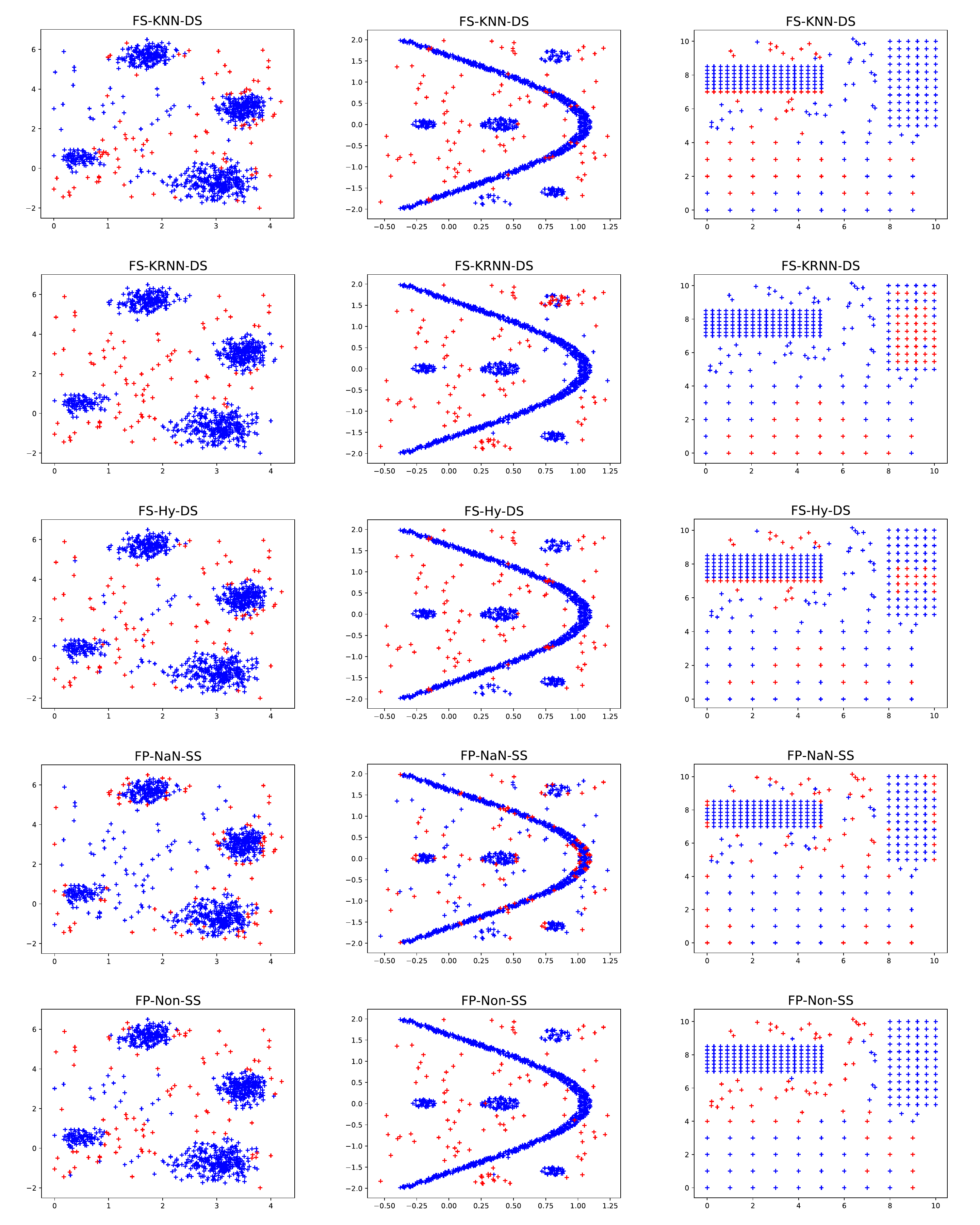}
	\caption{Outlier detection results of the DS-based methods at $K=20$.}
	\label{DSS}
\end{figure}

\begin{itemize}
	\item In Figure \ref{sss} and \ref{DSS}, the RKNN-based methods can hardly identify local outliers. It is noteworthy that Non-NaN based methods has the best result on Data3. This shows that the Non-NaN based methods may alleviate the problem of low density.
	\item Whether the neighbors are obtained by static sorting or dynamic selection, the NaN-based methods always maintains a low level. In Figure \ref{DSS}, NaN-based method identifies the points on the edge of cluster as outliers. This can be caused by the fact that the similarity of the data within the set is too high to distinguish between outliers and inliers. 
	\item RKNN-based methods can detection all outliers successfully on Data1 and Data2. Other methods often treat points on the boundary of clusters as outliers, this may leads a poor performance.
\end{itemize}

\subsection{Discussion}
As we can observe from the result of comparison experiments in Section \ref{4.4}, the performance of the algorithm can be further improved by reorganizing the three levels of components in the taxonomy. As shown in Table \ref{result1} and \ref{results2_1}, different neighbors favor different scenarios. For instance, RKNN-based methods lead to a significant increase in performance of KNN-based methods on Data1. However, they hardly works in data with low-density patterns. Furthermore, DS-based methods seem to be more suitable for working in the environment of high-dimensional data. They perform poorly on the subspace of the Wilt dataset besides synthetic datasets.

In Figure \ref{sss} and \ref{DSS}, most algorithms will recognize the samples in low-density areas as outliers and the Non-NaN based approaches is the most efficient in this scenario. We also investigated the effect of parameter K by visualization. In Figure \ref{resSS} and \ref{resDS}, different K value may lead to a substantial change in model decision. NaN-based method can be called the most unreliable method. Since all the objects in the NaN set are the most similar, it cannot bring benefits to distinguish outliers. In addition, the method based on static sorting has great advantages in time indicators. Therefore, they are suitable for real scenes.

\section{Conclusions} \label{sec5}
This paper introduces a taxonomy of neighbor-based methods for local outlier detection, which classifies existing algorithms from information, neighbor and methodology levels. Existing methods usually focus on designing different outlier factors, while ignoring the importance of neighbors. We discussed the advantages and disadvantages of various neighbors based on the proposed taxonomy. Experimental results verify that other neighbors can be used to replace KNN to further improve the algorithm. And the improvement can be achieved by a rational combination of components in the information, neighbor, and methodology levels.

Although satisfactory results have been achieved, there are still several problems to be further solved. On the one hand, the performance benefits brought by different subspaces need to be analyzed. On the other hand, it is crucial to design a method to find the best K value adaptively based on data. Comparative studies have shown that effective subspaces will bring benefits. We can find subspace by attribute reduction based on the complexity of each attribute of data. Furthermore, the method of selecting K value can be developed based on the stability of neighbor. Moreover,  some challenging tasks is valuable to explore, such as federated learning \cite{qi2022clustering,liu2023cross,qi2023cross,qiattentive} and visual classification \cite{chen2023class,chen2023class1,wang2023multi,wang2022meta,liu2022prompt,li2022unsupervised}.

\section*{Acknowledgments}
This work was supported by the National Natural Science Foundation of China (Grant No. 32270705)

\bibliographystyle{ws-ijwmip}
\bibliography{ref}






\end{document}